\documentclass{sig-alternate-05-2015}
\usepackage{flushend}

\makeatletter
\def\@copyrightspace{\relax}
\makeatother

\begin{document}

\setcopyright{acmcopyright}


\title{Depression Severity Estimation from Multiple Modalities}
%
%
%
%

\author{%
\begin{tabular}{ c c c }
  \begin{tabular}{c} Evgeny Stepanov \\ University of Trento, Italy \\ evgeny.stepanov@unitn.it \end{tabular} & 
  \begin{tabular}{c} Stephane Lathuiliere \\ INRIA Grenoble, France \\ stephane.lathuiliere@inria.fr \end{tabular} & 
  \begin{tabular}{c} Shammur Absar Chowdhury \\ University of Trento, Italy \\ shammur.chowdhury@unitn.it \end{tabular} \\
  \vspace{0.5mm} \\
  \begin{tabular}{c} Arindam Ghosh \\ University of Trento, Italy \\ arindam.ghosh@unitn.it \end{tabular} & 
  \begin{tabular}{c} Radu-Lauren\c{t}iu Vieriu \\ University of Trento, Italy \\ radulaurentiu.vieriu@unitn.it \end{tabular} & 
  \begin{tabular}{c} Nicu Sebe \\ University of Trento, Italy \\ niculae.sebe@unitn.it \end{tabular} \\
  \vspace{0.5mm} \\
  \begin{tabular}{c} \\ \\ \end{tabular} & 
  \begin{tabular}{c} Giuseppe Riccardi \\ University of Trento, Italy \\ giuseppe.riccardi@unitn.it \end{tabular} & 
  \begin{tabular}{c} \\ \\ \end{tabular} \\
\end{tabular}
}

%



\maketitle
\begin{abstract}
Depression is a major debilitating disorder which can affect people from all ages. With a continuous increase in the number of annual cases of depression, there is a need to develop automatic techniques for the detection of the presence and extent of depression. In this AVEC challenge we explore different modalities (speech, language and visual features extracted from face) to design and develop automatic methods for the detection of depression. In psychology literature, the PHQ-8 questionnaire is well established as a tool for measuring the severity of depression. In this paper we aim to automatically predict the PHQ-8 scores from features extracted from the different modalities. We show that visual features extracted from facial landmarks obtain the best performance in terms of estimating the PHQ-8 results with a mean absolute error (MAE) of 4.66 on the development set. Behavioral characteristics from speech provide an MAE of 4.73. Language features yield a slightly higher MAE of 5.17. When switching to the test set, our Turn Features derived from audio transcriptions achieve the best performance, scoring an MAE of 4.11 (corresponding to an RMSE of 4.94), which makes our system the winner of the AVEC 2017 depression sub-challenge.
\end{abstract}


\section{Introduction}
According to the World Health Organization (WHO), depression is a major mental disorder with about 300 million people of all ages affected worldwide. As per the  Global Burden of Disease Study ~\cite{ferrari2013burden}, depression is the second leading cause of disability worldwide and is on the rise. Depression affects every aspect of a person's life. People affected from depression often suffer from a certain extent of 
physical and social impairment. Side effects of depression include sleep disruptions or insomnia, drug or alcohol abuse, and overall loss of quality of life. If left untreated it can lead to complications such as reductions in the volume of the hippocampus \cite{sheline2003untreated}. Major clinical depression may even lead to suicide and annually the burden of death due to depression is on the rise. There is growing evidence that depression can cause impairment of the immune function by affecting different immunological pathways such as the central nervous system (CNS), the endocrine system, and the
cardiovascular system. This can lead to the development or aggravation of co-morbidities and worsen health conditions in other diseases. Nicholson et al \cite{nicholson2006depression}, through a meta-analysis of 54 cohort studies which performed follow up analysis of coronary heart diseases (CHD) showed that patients with major depression had an increased risk of developing fatal CHD.

Diagnosis of depression still remains a challenge.  Some symptoms of depression are not readily visible to others. Since depressed people often have decreased social contact, detection of the disease becomes difficult. Current diagnosis of depression is dependent on an evaluation by a psychiatrist supported by standard questionnaires to screen for depression. The Personal Health Questionnaire Depression Scale (PHQ-8) Scoring and the Hamilton Depression Rating Scale are two well established tools for the diagnosis of depression. However, these questionnaires need to be administered and interpreted by a therapist. The stigma around the disease and lack of understanding often prevents patients from seeking early psychiatric help. 

The growing burden of this disease suggests that there is a need to develop technologies which can aid in automatic detection and effective care of patients suffering from depression. Affective computing focuses on the sensing, detection, and interpretation of affective states of people from interactions with computers or machines. Research on affective computing uses modalities ranging from overt signals such as speech, language and video to covert signals such as heart rate, skin temperature, galvanic skin response to understand the mental and affective states of humans. While the initial goal of affective computing research was to build better computers which could understand and empathize with humans, the same techniques have been applied to turn computers into tools for automatically identifying psychological states and mental health. 

Therefore, the motivation of the study, is to explore different sources of information, such as audio, video, language and behavioral cues, to predict the severity of  depression. While doing so, we also investigate different feature representation and modeling techniques corresponding to each modality for improving the performance of automatic prediction.  

The paper is organized as follows. In Section \ref{sec:soa}, we present the literature review and the state of the art experiments performed for the detection of depression and affective disorders from speech, language, and facial expressions. This is followed by a brief description of the multi-modal data used for the study in Section \ref{sec:data}. An overview of the features and experimental methodology used in this study are given in Section \ref{sec:exp} and then we provide a conclusion in Section \ref{sec:concl}.

\section{State of the Art - Speech, language and facial expressions}
\label{sec:soa}
Speech, language and facial expressions are three of the major overt signals which have been widely used for interpreting human  psychological states. Automatic analysis of speech has been used for emotion recognition \cite{trigeorgis2016adieu, patel2017emotion}, stress detection \cite{hansen1996analysis, yogesh2017new}, and mood state characterisation \cite{vanello2012speech, chang2011s}. Natural language and speech processing from diaries and recordings have been used to detect the onset of dementia, alzheimers, and aphasia \cite{thomas2005automatic, fraser2013automatic}. Analysis of facial expressions have shown to be highly effective in tracking the progressive degeneration of cognitive health in patients suffering from schizophrenia and bipolar disorder \cite{bersani2013facial}.

\subsection{Speech and Language}

Several psychological conditions clearly manifest themselves through changes in speech patterns and language usage. Computational and automatic screening methods have the power  to detect micro-changes in speech and language patterns which would otherwise have gone unnoticed. Properties such as speech rate, pause duration and usage of fillers can be indicative of cognitive decline in individuals. Changes in prosody, and fluency can also be useful is detecting mental health changes of depressive patients. 

Research on the diagnosis of mental health from speech and language was pioneered by the German psychiatrist Zwirner \cite{zwirner1930beitrag} in early 1930. He designed a device capable of tracking fundamental frequency for the detection of mental health of patients suffering from depression. Newman and Mather \cite{newman1938analysis} in 1938 carried out similar experiments to systematically record  patient's speech as they read pre-defined text and interacted with a psychiatrist. This data was analysed to show that there were distinct speech features such as speech tempo, prosodic pauses, absence of glottal rasping associated with patients suffering from affective disorders.

France et. al \cite{france2000acoustical} performed multivariate feature and discriminant analyses on the speech data from 67 male and 48 female subjects to show that formant and power spectral density (PSD) based features demonstrated the highest discriminative powers for classification in both genders. Pope et al \cite{pope1970anxiety} investigated the relationship between anxiety and depression and speech patterns to show that anxiety was positively correlated with speech disturbances and resistivity in speech. He also found that silent pauses were positively correlated with depression \cite{pope1970anxiety}.

Kaya et al \cite{kaya2014cca} demonstrated that feature selection techniques based on canonical correlation analysis (CCA) can be effective in detecting depression from speech signals.

Wang et al \cite{wang2013depression} applied data mining techniques to build models which achieved a precision of 80\% for detecting depression based on sentiment analysis of users on a Chinese micro-blogging platform. Rumshisky et al \cite{rumshisky2016predicting} demonstrated through a study on 4687 patients that NLP techniques such as topic modeling can be used to improve prediction of psychiatric readmission.

\subsection{Face Analysis}

Facial expressions can be an extremely powerful medium used to convey human overt emotional feedback. In recent times, there has been significant progress in developing methods for facial feature tracking for the analysis of facial expressions and the detection of emotions. Studies have shown that it is possible to effectively detect the presence of pain shown on faces.

Machine learning techniques have been shown to be effective for the automatic detection of pain and mental state from facial expressions. Littleworth et al. \cite{littlewort2009automatic} used a two-stage system to train machine learning algorithms to detect expressions of real and fake pain. Their classifier obtained an accuracy of 88\% compared to an accuracy of 49\% demonstrated by naive human subjects used in their study. Ambadar et al \cite{ambadar2009all} demonstrated that analysis of facial expression can be used to classify smiles into three distinct categories - amused, polite and nervous.

One of the most popular technique used for capturing the subtlety and fine-grained variations in facial expression is the Facial Action Coding System (FACS) developed by Ekman and Freisen. The FACS is based on the consensus of the judgment human experts who observe pre-recorded facial expressions and perform manual annotation of FACS codes for each frame. These annotations, which are called action units (AUs) can belong to one of 44 different classes. FACS has been widely used in the field of psychology for measuring emotions, affect, and behavior \cite{craig2008emote, bartlett2006fully,russell1994there}. More recently \cite{girard2014nonverbal}, FACS have been shown to be correlated with depression severity. Specifically, \cite{girard2014nonverbal} found that severely depressed subjects are more likely to show fewer affiliative facial action units (AU12 and AU15) and more non-affiliative ones (AU14). 

Head pose and eye gaze have also been shown to encode information about depression. For instance \cite{girard2014nonverbal} observes that an increase in the severity of depression comes with a diminished head motion. Other works \cite{alghowinem2013head,joshi2013can,scherer2014automatic} have also investigated the link between head pose, eye gaze and depression, all evidence that such a link exists and it is all worth considering.

\subsection{Combination}
Combination of facial expressions, speech and multimodal information can be used to enhance the recognition of human mental state. Busso et al. \cite{busso2004analysis} demonstrated that both feature fusion (early fusion) and decision fusion (late fusion) from the different modalities outperformed individual features-based classification.

Dibeklioglu et al \cite{dibekliouglu2015multimodal} combined speech, facial movement and head movement to achieve an accuracy of 88.9\% for the detection of depression from clinical interviews. The accuracy of the combined signal streams exceeded the accuracy of single modalities to show that multimodal measures can be powerful for detection of depression. Alghowinem et al \cite{alghowinem2016multimodal} also demonstrated similar findings in their research to show that a combination of head pose, eye gaze and paralinguistic features yielded better performance than unimodal schemes.

\section {AVEC Audio Video Database}
\label{sec:data}
The 2017 Audio/Video Emotion Challenge and Workshop (AVEC 2017) ``Real-life depression" provides a corpus comprising of audio and video recordings and transcribed speech from the Distress Analysis Interview Corpus (DAIC) \cite{gratch2014distress}.

The dataset comprises of recordings from 189 sessions of human agent interaction where each subject was interviewed by a virtual psychologist (see Table \ref{tbl:data} for the distribution of labels in the training and development sets). The audio files, transcripts and continuous facial features of the human subject is provided as part of the challenge. The Personal Health Questionnaire Depression Scale (PHQ-8) score of the subjects is also provided in the dataset. The PHQ-8 \cite{kroenke2009phq} is a set of 8 short multiple choice questions which has been established as a diagnostic tool for the measurement of the severity of depressive disorders. Automatic estimation of the PHQ-8 score from different modalities such as speech and video can aid in the early detection of depression and monitoring of depressive states. In the AVEC challenge, the goal is to look at different streams of data recorded during a session with the subject to predict the PHQ-8 scores, and to classify the subject as depressed or not.

\begin{table}
\centering
\caption{Distribution of the AVEC data set into training and development sets for depressed (\textbf{D}) and non-depressed (\textbf{ND}) classes, and overall (\textbf{ALL}).}
\label{tbl:data}
\begin{tabular}{|l|rr|rr|r|}
\hline
& \multicolumn{2}{c|}{\textbf{ND}}
& \multicolumn{2}{c|}{\textbf{D}}
& \multicolumn{1}{c|}{\textbf{ALL}}\\
\hline
\textit{Training}     & 77 & (72\%) & 30 & (28\%) & 107 \\
\textit{Developement} & 23 & (66\%) & 13 & (34\%) & 35 \\
\hline
\end{tabular}
\end{table}

\section{Experiments}
\label{sec:exp}
In this section we describe the experiments conducted for the feature extraction and regression experiments conducted on the speech, behavioral, language and 
\subsection{Speech and Behavioral Characteristic Features}
\subsubsection{Acoustic Features}
To understand the predictive characteristics of low-level acoustic feature groups to assess the depression severity of the participant, we extracted low-level descriptors (LLDs) from the participant's turns in each conversation. For this, we have extracted different groups of low-level features using openSMILE \cite{eyben2013recent}, motivated by their successful utilization in several paralinguistic tasks \cite{schuller2011recognising,alam2013comparative,shammur2017overlap,chowdhury2015overlapiccasp}. These sets of acoustic features were extracted with approximately 100 overlapping frames per second and with 25 milliseconds of window. The low-level features are extracted as three groups including: 
\begin{itemize}
\item Spectral features (\textbf{S}) such as energy in spectral bands (0-250Hz, 0-650Hz, 250-650Hz, 1-4kHz), roll-off points (25\%, 50\%, 70\%, 90\%), centroid, flux, max-position and min-position.
\item Prosodic features (\textbf{P}) such as pitch (Fundamental frequency f0, f0-envelope), loudness, voice-probability.
\item Voice Quality features (\textbf{VQ}) such as jitter, shimmer, logarithmic harmonics-to-noise ratio (logHNR).
\end{itemize}

\noindent These low-level features are then projected on 24 statistical functionals, which include range, absolute position of max and min, linear and quadratic regression coefficients and their corresponding approximation errors, zero crossing rate, peaks, mean peak distance, mean peak, geometric mean of non-zero values, number of non-zeros and moments- centroid, variance, standard deviation, skewness, and kurtosis.

\subsubsection{Behavioral Characteristics Features}
Apart from extracting low-level features from raw speech signals, we also explored the transcription.

We crafted features that can capture information regarding the participant's non-vocal behavior (NB) along with their turn-taking behaviors (TB) and participants' Previous Diagnosed Information (PDI) features. 
The non-vocal behavior ($|NB|=3$) includes: 
\begin{itemize}
\item frequency of laughter in participant's turns.
\item percentage of disfluencies in the participant's turns, which might indicate hesitations.
\item counts of cues that might suggest inconvenience like whistling, mumbling, whispering or taking deep breaths among others.
\end{itemize}

The features that are used to describe the turn-taking behaviors, ($|TB|=6$) are the first and third quartiles and the median duration of respond time (in seconds) of the participants. Similarly we also extracted statistics for the with-in speaker silence (pause). The respond time represents how long the participants took to respond to the previous turn of the agent. 

The PDI feature set ($|PDI|=3$) contained numerical representations of the response of the participants to queries such as having any Post-traumatic Stress Disorder (PTSD), {\b ptsd}, depression {\b dep}, even having any military backgrounds {\b mb}. Each individual feature is encoded into three values (-1,0,1) where -1 represents the query is not present in the session, 0 presents a disconfirmation (e.g ptsd=0 means the participant responded as ``no'' to the previous turn query) and 1 presents confirmation of the query.

\subsubsection{Methodology and Results}
For the regression task, we studied the performance of acoustic and behavioral characteristics features. For modeling individual acoustic feature groups and their linear combination we used support vector machine for regression, implemented in weka \cite{hall2009weka} using Radial Basis Function (RBF) kernel with $\gamma=0.01$ and $C=1.0$. 

As for the linear combination of different acoustic feature groups, we first merged all the feature vectors linearly to obtain vector $M$, as shown in Equation \ref{eq:m}
\begin{eqnarray}
\label{eq:m}
\textbf{M}=P\cup S \cup VQ = \left \{ p_{1},.., p_{m}, s_{1}..,s_{n},v_{1},..,v_{l}  \right \}
\end{eqnarray}
where feature vectors P, S and VQ stands for prosody, spectral and voice quality as presented in Equations \ref{eq:p}-\ref{eq:v}.
\begin{eqnarray}
\label{eq:p}
P=\left \{ p_{1}, p_{2}, ..., p_{m}  \right \} \\
\label{eq:s}
S=\left \{ s_{1}, s_{2}, ..., s_{n}  \right \} \\
\label{eq:v}
VQ=\left \{ v_{1}, v_{2}, ..., v_{l}  \right \} 
\end{eqnarray}

From the merged feature vector we selected relevant feature subset $Fs-M$ using training set only. For the automatic feature selection, we used Relief feature selection technique \cite{kononenko1994estimating,Robnik-Sikonja1997}, successfully used in paralinguistic tasks \cite{alam2013comparative,chowdhury2014overlap}. The technique calculates the weight of the features based on the nearest $k$ instances ($k=20$, used for this study) of the same and different classes to rank each features.
Then by using a threshold, $th=0.02$, we selected top 20 features to use for the regression task. These parameters ($th$={0.02, 0, -0.02} and $k$={5,10,15,20}) are tuned using 3-fold cross validation of the training set.

As for predictor using behavioral characteristic feature group, we used Reduced Error Pruning Tree (``REPT'') implemented in weka \cite{hall2009weka}, which is a fast regression tree learner that uses information of variance reduction and prunes it using reduced error pruning.

\begin{table}[]
\centering
\caption{Results of individual acoustic feature groups with linearly merged feature groups and with Relief feature selection for depression severity estimation on the development set. $\star$ represents results tuned using 3-fold cross validation on the training set. $|F|$ represent feature set dimension.}
\label{rslt-ac}
\scalebox{1.0}{
\begin{tabular}{|c|c|c|c|}

\hline
\textbf{Feature set, F} & \textbf{|F|} & \textbf{RMSE} & \textbf{MAE} \\ \hline\hline
\textit{Spectral} & 864 & \textbf{6.32} & \textbf{4.96} \\ 
\textit{Voice Quality} & 288 & 7.05 & 5.70 \\
\textit{Prosody} & 288 & 7.10 & 5.75 \\ \hline\hline
\textit{Merged} & 1440 & 6.43 & 5.40 \\ \hline\hline
\textit{Merged+Feat.Selection}$\star$ & 20 & 6.70 & 5.20 \\ \hline
\end{tabular}}
\end{table}

\begin{table}[]
\centering
\caption{Result for depression severity estimation using behavioral characteristic features on development set. $|F|$ represent feature set dimension.}
\label{rslt-bc}
\scalebox{1.0}{
\begin{tabular}{|c|c|c|c|}

\hline
\textbf{Feature set, F} & \textbf{|F|} & \textbf{RMSE} & \textbf{MAE} \\ \hline\hline
\textit{Behavioral characteristic} & 12 & \textbf{\textit{5.54}} & \textbf{\textit{4.73}} \\ \hline
\end{tabular}}
\end{table}

The results are presented in Table \ref{rslt-ac} for individual feature set and their combinations. The result indicated that spectral features are a good predictor of PHQ score compared to all other settings presented in the table. 
It is observed that even feature selection on the merged vector also performed better than other sets except spectral and is above the baseline, i.e., $MAE=5.36$ and $RMSE=6.74$ on the same development set. The selected features include features from spectral group (75\%), prosodic group (20\%) and voice quality (5\%) group.

It is also observed that using behavioral characteristic features, we obtained a decrease of both MAE and RMSE by a magnitude of $0.63$ and $1.20$ respectively compared to all the results reported in the AVEC2017 baseline manuscript. Further analysis using feature ranking technique, Relief, indicated that the PDI features especially \textit{dep} and \textit{ptsd} are the top ranked features followed by the median of the response time, the quartiles of the within-speaker silence duration and laughter frequency.

\subsection{Language}

Additional to the speech-based features, we explore text-based representations to predict depression severity estimates. The widely used representation of a document in NLP is bag-of-words, where a document is represented by word occurences ignoring the order in which they appear. We experiment both with binary (BOOL) and tf-idf (TFIDF) weighted representations. While the binary representation encodes words that are present in the document regardless of their frequency, tf-idf weighted representation considers both the frequency of the term (\textit{tf}) in a document and the inverse document frequency (\textit{idf}) -- which lowers the weight of the very frequent terms in a collection and increases the weight of the rare terms with respect to the equations \ref{eq:tfidf}-\ref{eq:idf}.

\begin{eqnarray}
tf-idf(t,d) & = & tf(t,d) * idf(t) \label{eq:tfidf}\\
idf(t) & = & log \frac{n_d}{df(d,t)} + 1 \label{eq:idf}
\end{eqnarray}

\noindent Where $tf(t,d)$ is the term frequency, $n_d$ is the total number of documents, and $df(d,t)$ is the frequency of documents containing the term.

Besides bag-of-words representation, we also experiment with the word embedding representation (WE) \cite{mikolov2013efficient}, where pre-trained per-word embedding vectors are averaged for a document. We make use of the SKIPGRAM embedding vectors pre-trained on GoogleNews with a embedding dimension 300 and window 10.

Since the provided speech transcripts are of human-machine conversations, we first extract human turns and convert them into bag-of-words representation. The transcripts contain annotations for the speech phenomena such as laughter, sigh, etc., which were treated as any other token. Thus, the representation implicitly encodes the presence of these phenomena in the conversation; and also its frequency in the case of tf-idf based representations. For the word embedding representation, however, this is not the case, as there are no pre-trained vectors for these. 

\begin{table}
\centering
\caption{Root mean square error (RMSE) and mean absolute error (MAE) for depression severity regression using lexical features and Support Vector regression with linear kernel on the development set for the mean baseline (\textit{BL: mean}), binary (\textit{BOOL}), tf-idf weighted (\textit{TFIDF}) bag-of-words representations, and averaged word embedding vectors (\textit{WE}). We also provide the audio and audio-video feature-based baselines (\textit{BL: Audio} and \textit{BL: Audio-Video}) using Random Forests.}
\label{tbl:txt}
\begin{tabular}{|l|c|c|}
\hline
& \textbf{RMSE} & \textbf{MAE}\\
\hline
\textit{BL: mean}        & 6.57 & 5.50 \\
\textit{BL: Audio}       & 6.74 & 5.36 \\
\textit{BL: Audio-Video} & 6.62 & 5.52 \\
\hline
\textit{BOOL}     & \textbf{6.31} & \textbf{5.17} \\
\textit{TFIDF}    & 6.78 & 5.40 \\
\textit{WE}       & 6.84 & 5.41 \\
\hline
\end{tabular}
\end{table}

The algorithm of our choice for text-based representations is Support Vector Regression (SVR) with linear kernel, implemented in scikit-learn \cite{scikit-learn}. The regression results for each of the document representations are given in Table \ref{tbl:txt} in terms of RMSE and MAE. We also provide a mean baseline (BL:mean) and the audio and audio-video feature-based baselines\footnote{Cite the baseline paper}. As it can be observed, the only representation that outperforms all the baselines is the binary bag-of-word representation that yields RMSE=6.31 and MAE=5.17.

\subsection{Visual Features}
Inspired by \cite{valstar2016avec} and the success reported in \cite{yang2016decision}, we use the 68 3D facial keypoints and compute geometric features as follows: for every facial representation, we first remove the 3D bias (equal to a translation in the Euclidean space by subtracting the mean value in 3D), then we normalize the resulting representation so that the average distance to the center (origin) is equal to 1. Finally, we compute Euclidean distances between all possible pairs of 3D normalized points and add them to the normalized representation. This results in a feature vector of size $2482$. Consequently, we reduce this dimension by applying PCA and keeping over 99.5\% of variance, resulting in a feature vector of size 33.

Since we are dealing with video sequences, we propose to regress depression using models naturally designed for temporal data. Specifically, we propose the use of LSTMs \cite{hochreiter1997long} for this task. LSTMs have emerged as an effective and scalable model for several learning problems related to sequential data, such as handwriting recognition \cite{pham2014dropout,doetsch2014fast}, generation of handwritten characters \cite{graves2013speech}, language modeling and translation \cite{zaremba2014recurrent,luong2014addressing}, audio \cite{marchi2014multi} and video \cite{donahue2015long} signal analysis, acoustic speech modeling \cite{sak2014long} and others. They have proved effective at capturing long-term temporal dependencies without suffering from the optimization hurdles that plague simple recurrent neural networks (RNNs).

In order to build our training set, we apply a sliding window approach to the video sequences, using windows of size $W$, overlapped by $O$ samples. We use the \textit{success} flag provided by the dataset creators which models the tracking confidence for each frame. We adopt a 0-tolerance strategy and discard all windows for which at least one failed tracking is present. We do this to exclude the risk of introducing artifacts into the feature space, that the model might misleadingly exploit for solving the task. We set the values for $W$ and $O$ empirically to $60$ and $30$, respectively. We downsample the data to 1 second, which makes our windows 1 minute long, with an overlap of 30 seconds. During testing, we apply the same window-ing scheme and average the window-level predictions over the length of the test sequence.

Next, we train a double layered LSTM model on regressing depression at window level on the training set. The model is composed of two stacked layers of size 16, followed by a $Dense$ layer with a $linear$ activation function. We use dropout \cite{srivastava2014dropout} equal to $0.5$ to control overfitting and batch normalization \cite{ioffe2015batch} to limit internal covariance shift. As loss function, we use the mean squared error. In order to validate our LSTM model, we perform a leave-one-sequence-out cross-validation scheme on the training set. After 100 epochs, our models achieve an MAE of 4.97 and an RMSE of 6.26, which we find encouraging. We further retrain the model on the full training set and monitor the performance on the development partition.

\begin{figure}[h!]
\centering
\includegraphics[scale=0.4]{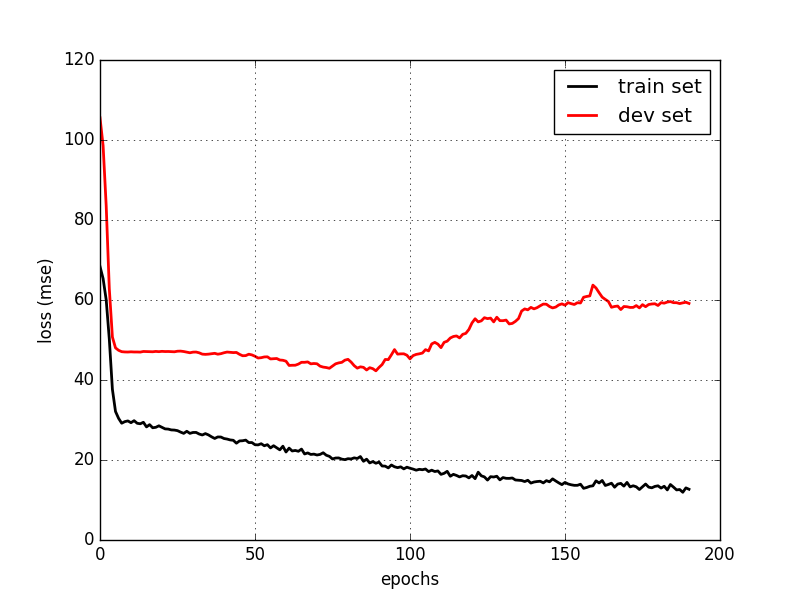}
\caption{LSTM learning curves: trainset (black) and development set (red). We note the existence of a turning point in the validation loss, typically used as a good compromise between underfitting and overfitting}
\label{fig:learning_curves}
\end{figure}

Figure \ref{fig:learning_curves} shows the learning plots of the loss function during training for both training (black) and validation (red) sets. We observe a monotonic decrease of the loss function on the training set, while on the validation, the behavior is a typical decrease, followed by an increase of the same loss. We use the validation set to early stop the training, thus resulting in a model ($lstm\_opt$) with the best performance on this set. 

Following the baseline manuscript, we report in Table \ref{tab:vision_results} as performance measures the RMSE and MAE of $lstm\_opt$ on both train as well as test sets. In addition to the requested quantities, we also report the explained variation regression score (EVS), defined as:

\begin{equation}
    evs(y,\widehat{y})=1-\frac{Var(y-\widehat{y})}{Var(y)}
    \label{eq:evs}
\end{equation}

where $Var$ represents the statistical variance. EVS measures the degree to which a model (in our case $lstm\_opt$) accounts for the variance of a given set of labels through the predictions it makes. The upper bound of EVS is 1 and corresponds to a perfect modeling.

\begin{table}[]
\centering
\caption{Performance measures obtained using our LSTM model on the training set as well as on the development set. We also report the explained variance regression score (EVS), which measures the degree to which the model "explains" the variation of the ground truth labels using the predictions (see Equation \ref{eq:evs} for a formal definition)}
\label{tab:vision_results}
\begin{tabular}{|l|l|l|l|} \hline
          & RMSE & MAE  & EVS  \\ \hline
train set & 3.17 & 2.32 & 0.66 \\ \hline
dev set   & 6.09 & 4.66 & 0.15 \\ \hline
\end{tabular}
\end{table}

As can be observed from Table \ref{tab:vision_results}, our LSTM model fits well the training set and manages to score a promising MAE on the development partition, better than all reported values in the AVEC2017 baseline manuscript as well as in the last year's winning paper \cite{yang2016decision}. 

\subsection{Results on the test set}
We submitted four trials for evaluation on the held out test set. Results are depicted in Tab. \ref{tab:test_results}. The behavioral characteristic features extracted from audio transcriptions achieve the lowest errors on the test partition, which is unsurprising considering the promising cross-validation results obtained on the the development set (\textit{i.e.} RMSE of 5.54 and MAE of 4.73). What is slightly surprising though is the performance of the visual features. Despite achieving an encouraging MAE on the development set, our LSTM model failed to generalize well enough to unseen data.

\begin{table}[]
\centering
\caption{Results on the test set}
\label{tab:test_results}
\begin{tabular}{|c|c|c|} \hline
                           & RMSE & MAE  \\ \hline
Spectral features (speech) & 6.63 & 5.08 \\ \hline
Turn features (speech)     & 4.94 & 4.11 \\ \hline
Text features              & 5.83 & 4.88 \\ \hline
Video features 			   & 6.72 & 5.36 \\ \hline
\end{tabular}
\end{table}

\section{Conclusions}
\label{sec:concl}
In this paper we address the depression sub-challenge problem formulated in AVEC2017, \textit{i.e.} regressing PHQ-8 depression scores from multi-modal data. We process different modalities (audio, language, visual) accompanying the corpus and developed regression systems separately. In the audio domain, we find the spectral features to be most suited for this task, achieving an MAE score of 4.96 on the development set (RMSE = 6.32) while lexical features score no lower than 5.17 (MAE) and 6.31 (RMSE). Despite being the worst performing modality in the baseline manuscript, visual features achieve the smallest errors on the development set in our experiments. Using a sliding window approach and temporal modeling, we obtain an MAE of 4.66 (RMSE = 6.09). We also observed that behavioral cues extracted from transcripts achieve smaller errors (MAE = 4.73, RMSE = 5.54) compared to audio and language features and are good predictor of the depression severity scores. When studied further, we found that previous diagnosed information cues, participants' response time to the agent among others are one of the most informed feature to predict the depression PHQ-8 scores. This is indeed confirmed by the results obtained on the test set, where behavioral cues scored the smallest MAE values among all other feature sets.

In this paper, we have studied each modality individually to understand its strength in estimating the depression severity. In future work, we plan investigating how we can combine individual modalities to improve the overall performance.




\bibliographystyle{abbrv}
\bibliography{sigproc}  
\end{document}